\documentclass[letterpaper, 10 pt, journal, twoside]{IEEEtran} 

\IEEEoverridecommandlockouts                              

\usepackage{graphics} 
\usepackage{epsfig} 
\usepackage{times} 
\usepackage{amsmath} 
\usepackage{amssymb}  
\usepackage{breqn}
\usepackage{bm}
\usepackage{algorithm}
\usepackage{algpseudocode}
\usepackage{physics}
\usepackage[export]{adjustbox}
\usepackage{caption}
\usepackage{mathtools}
\usepackage{xcolor}
\usepackage{hyperref}

\newcommand{\bs}{\boldsymbol}

\newcommand{\xii}{\bs{\xi}}
\newcommand{\gammai}{\bs{\gamma}}
\newcommand{\Lambdai}{\bs{\Lambda}}
\newcommand{\Omegai}{\bs{\Omega}}
\newcommand{\Phii}{\bs{\Phi}}
\newcommand{\varpidot}{\dot{\bs{\varpi}}}

\newcommand{\varpii}{\bs{\varpi}}
\newcommand{\xidot}{\dot{\bs{\xi}}}
\newcommand{\xiddot}{\ddot{\bs{\xi}}}
\newcommand{\jac}{\bs{\mathcal{J}}}

\newcommand{\bcf}{\;\mbox{\boldmath ${\cal F}$\unboldmath}}
\newcommand{\argminD}{\arg\,\min}

\def\Vec#1{\!\!\hbox{$#1$\kern-0.38em\lower0.85em\hbox{$\vec{}\,$}}\,}%

\title{A White-Noise-On-Jerk Motion Prior for Continuous-Time Trajectory Estimation on \textit{SE(3)}}

\author{Tim Y. Tang$^{1}$, David J. Yoon$^{1}$, and Timothy D. Barfoot$^{1}$%
\thanks{Manuscript received: September 10, 2018; Revised December 2, 2018; Accepted December 21, 2018.}
\thanks{This paper was recommended for publication by Editor Cyrill Stachniss upon evaluation of the Associate Editor and Reviewers' comments. 
This work was supported by Applanix Corporation and the Natural Sciences and Engineering Research Council of Canada (NSERC). Our test vehicle was donated by General Motors (GM) Canada.} 
\thanks{$^{1}$The authors are with the Autonomous Space Robotics Lab at University of Toronto, Toronto, Ontario, Canada.
        {\tt\small \{tim.tang, david.yoon\}@robotics.utias.utoronto.ca}, {\tt\small tim.barfoot@utoronto.ca}}
\thanks{Digital Object Identifier (DOI): see top of this page.}
}

\begin{document}
\maketitle

\begin{abstract}
Simultaneous trajectory estimation and mapping (STEAM) offers an efficient approach to continuous-time trajectory estimation, by representing the trajectory as a Gaussian process (GP). Previous formulations of the STEAM framework use a GP prior that assumes white-noise-on-acceleration, with the prior mean encouraging constant body-centric velocity. We show that such a prior cannot sufficiently represent trajectory sections with non-zero acceleration, resulting in a bias to the posterior estimates.

This paper derives a novel motion prior that assumes white-noise-on-jerk, where the prior mean encourages constant body-centric acceleration. With the new prior, we formulate a variation of STEAM that estimates the pose, body-centric velocity, and body-centric acceleration. By evaluating across several datasets, we show that the new prior greatly outperforms the white-noise-on-acceleration prior in terms of solution accuracy.
\end{abstract}

\begin{IEEEkeywords}
SLAM, Localization
\end{IEEEkeywords}

\section{Introduction}  
\IEEEPARstart{S}{tate} estimation techniques for mobile robotics have been predominantly formulated in discrete time. While discrete-time techniques are sufficient for many applications, they are not ideal for high-rate sensors that take measurements continuously along a trajectory (e.g., scanning-while-moving lidars), or a combination of asynchronous sensors. Continuous-time estimation techniques are much more suitable in these cases, since measurements can be incorporated at any time along the trajectory, without needing to include an additional state at every measurement time. Moreover, continuous-time techniques have the advantage that the posterior estimates can be queried at any time along the trajectory, not just at measurement times.

\begin{figure}[!htbp]
\centering
\includegraphics[height=2.09in]{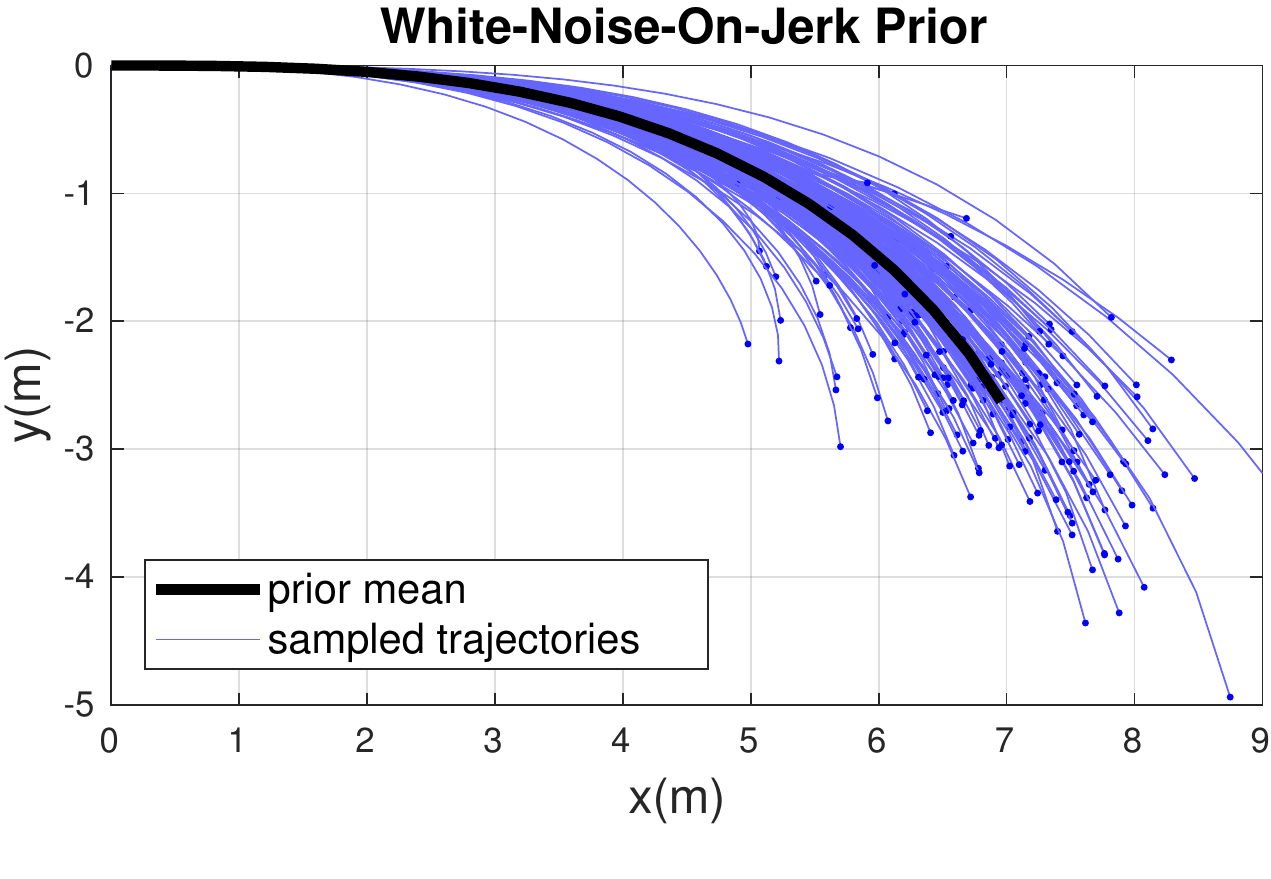}
\includegraphics[height=1.95in]{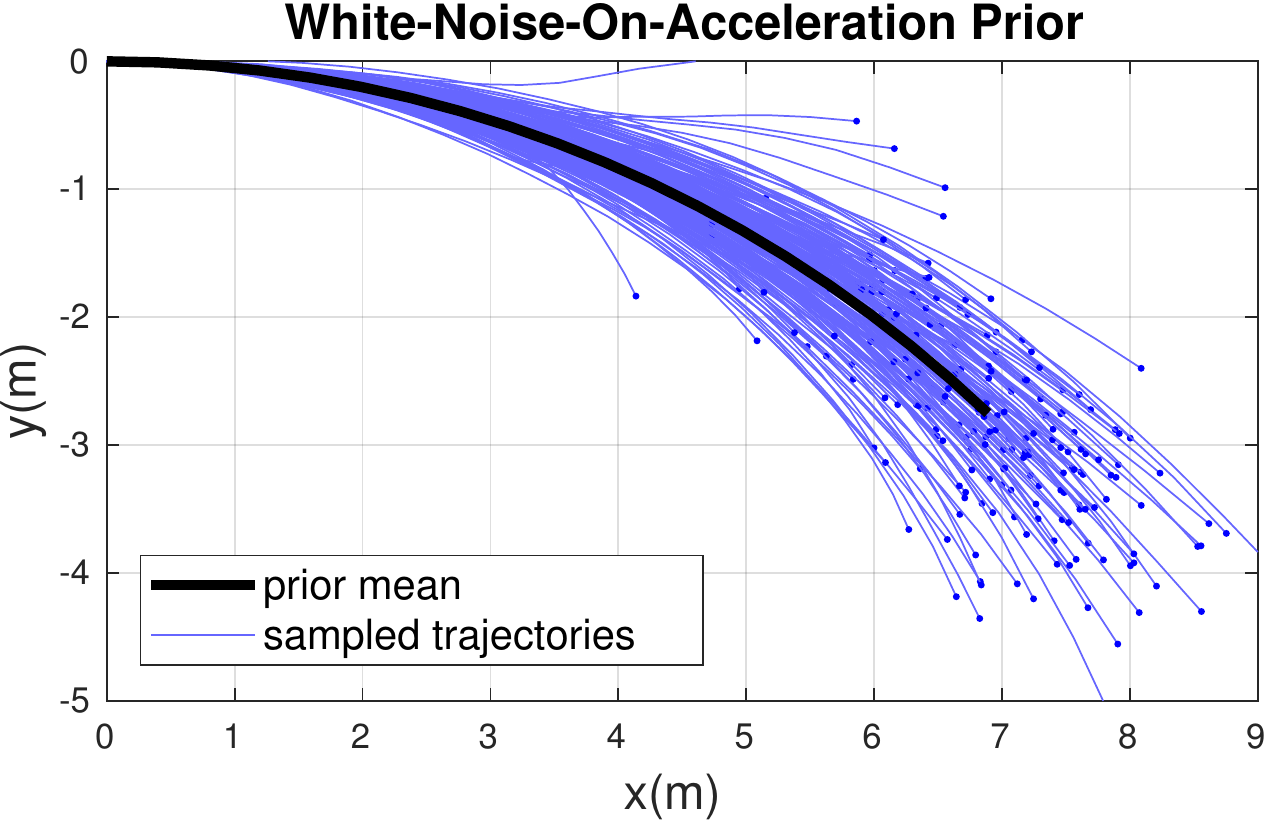}
\captionof{figure}{\footnotesize \label{fig:traj_samples} Existing formulations of STEAM use a white-noise-on-acceleration motion prior (bottom), which have trouble representing trajectories with non-zero acceleration, such as in the motion of a vehicle in urban driving. We propose a white-noise-on-jerk motion prior (top), which is more suitable for representing these types of trajectories.}
\vspace{-5mm}
\end{figure}


Continuous-time estimation techniques can be categorized into two types: parametric and nonparametric. Parametric approaches typically represent the trajectory using a finite set of temporal basis functions. Our work focuses on the nonparametric approach in which the trajectory is represented as a Gaussian process, with time as the only input variable. While model fidelity in parametric approaches are affected by choices regarding trajectory representation and discretization, the GP approach relies heavily on the continuous-time prior distribution for solution quality.

Current formulations of the GP approach to continuous-time trajectory estimation employ a white-noise-on-acceleration (WNOA) prior, or one that assumes the prior mean is constant-velocity. While this choice of prior is appropriate for certain types of motion, we argue that it is insufficient for representing trajectories with non-zero acceleration, such as in the motion of a vehicle in urban driving. We show that a bias can occur when the motion prior does not sufficiently represent the underlying trajectory.

With this in mind, we derive a white-noise-on-jerk (WNOJ) motion prior, which assumes the prior mean is constant-acceleration. Our derivation starts with the same form of physically motivated stochastic differential equation (SDE) for describing motion as in the WNOA prior.

By evaluating on several real-world lidar datasets, we show that our variation of STEAM with the WNOJ prior greatly outperforms the current formulation of STEAM, which employs a WNOA prior. In particular, the use of WNOJ prior results in reduced bias and improved odometry accuracy to the estimated trajectory. We perform the experimental evaluation using lidar-only motion estimation, as this is a problem particularly suitable for continuous-time methods. The contribution of this paper, however, can be applied to any choice of sensor suite.

In Section \ref{sec:related_work} we review previous work. An overview of our existing continuous-time lidar-only estimator is provided in Section \ref{sec:ct_estimator}. In Section \ref{sec:estimator_bias} we identify a source of estimator bias that relates to the choice of GP prior. Section \ref{sec:derivation} presents the derivation of a white-noise-on-jerk motion prior, which is compared against the white-noise-on-acceleration prior experimentally in Section \ref{sec:experimental_validation}. In Section \ref{sec:conclusion} we give concluding remarks and discuss future work.

\section{Related Work}
\label{sec:related_work}
Early works on continuous-time estimation are mostly parametric approaches, which represent the trajectory using temporal basis functions. Jung and Taylor \cite{jung2001camera} first presented an estimator where the sensor trajectory is modelled by spline functions. Furgale et al. \cite{furgale2012continuous} derived the simultaneous localization and mapping (SLAM) problem in continuous time, and showed a small number of basis functions can sufficiently represent the state. Anderson and Barfoot \cite{anderson2013towards} derived a relative coordinate formulation by estimating the body-centric velocity. Lovegrove et al. \cite{lovegrove2013spline} applied continuous-time estimation in visual-inertial SLAM.  Recent work by Dub\'{e} et al. \cite{dube2016non} explored strategies for selecting knot sampling in parametric approaches to continuous-time trajectories.


Batch nonparametric approaches, which represent the trajectory as a Gaussian process, were first formulated by Tong et al. \cite{tong2013gaussian}. The smoothness assumption is handled in a principled manner through the underlying GP prior. Barfoot et al. \cite{barfoot2014batch} extended the GP approach to STEAM, which employs a WNOA motion prior, by jointly estimating the pose and velocity. This choice of GP motion prior results in the inverse kernel matrix being exactly sparse, leading to a very efficient formulation. Anderson and Barfoot  \cite{anderson2015full} extended \cite{barfoot2014batch} to matrix Lie groups. Boots et al. \cite{yan2017incremental} re-formulated STEAM from batch estimation to an incremental algorithm. STEAM has been applied to motion planning \cite{mukadam2017simultaneous}, crop monitoring \cite{dong20174d}, and visual teach and repeat \cite{warren2018towards}.

Without explicitly treating the trajectory as a continuous function of time, some estimators use interpolation between discrete poses to compensate for motion distortion, particularly in the case of scanning lidars. Bosse and Zlot \cite{bosse2009continuous}, \cite{zlot2014efficient} used cubic splines to enforce smoothness for a trajectory estimated using data from a 2D spinning lidar, and linearly interpolated between sampled poses. Dong et al. \cite{dong2014lighting} performed visual odometry from lidar intensity images, and interpolated on rotation and translation using a scheme detailed in \cite{barfoot2017state}. State-of-the-art lidar-only motion estimation algorithm, LOAM \cite{zhang2014loam}, interpolates on $SE(3)$ between adjacent discrete poses, using a scheme similar to \cite{bosse2009continuous}. Unlike continuous-time methods, these methods need to make ad-hoc assumptions about trajectory smoothness in order to carry out interpolation.

While various motion estimation methods have made assumptions with the trajectory being constant-velocity \cite{davison2007monoslam}, \cite{barfoot2014batch}, \cite{anderson2015full}, \cite{anderson2013ransac}, \cite{hedborg2012rolling}, a constant-acceleration trajectory assumption has been used for tracking control \cite{jang2015trajectory},  manipulator motion planning \cite{mukadam2017continuous}, and manipulator state estimation \cite{olofsson2016sensor}. To the best of our knowledge, the derivation we present in this paper is the first attempt at modelling the trajectory as constant-acceleration mean (white-noise-on-jerk) in the context of continuous-time trajectory estimation on $SE(3).$

\section{Continuous-Time Estimator}
\label{sec:ct_estimator}
In this section, we give details on our existing continuous-time lidar odometry algorithm, which uses the STEAM framework with a WNOA motion prior \cite{anderson2015full}. This serves as the baseline against which the WNOJ prior will be evaluated.

\subsection{WNOA GP Prior}
Our goal is to employ a class of GP priors that leads to an efficient formulation and a simple solution \cite{barfoot2014batch} \cite{barfoot2017state}. This class of GP priors is based on linear time-invariant (LTI) stochastic differential equations (SDEs) of the form
\begin{equation}
\label{eq:lti_sde}
\begin{split}
\dot{\gammai}(t) &= \bs{A}\gammai(t) + \bs{B}\mathbf{u}(t) + \bs{L}\mathbf{w}(t), \\
\mathbf{w}(t) &\sim \mathcal{GP}(\mathbf{0}, \mathbf{Q}_c \delta(t-t')),
\end{split}
\end{equation}
where $\gammai(t)$ is the state at timestep $t,$ $\mathbf{u}(t)$ is an exogenous input, and $\mathbf{w}(t)$ is a zero-mean, white-noise GP with \textit{power spectral density matrix}, $\mathbf{Q}_c\in \mathbb{R}^{6\times6}\ \cite{barfoot2017state}.$ If $\mathbf{u}(t)= \mathbf{0}, $ then for the mean function we have the simple solution
\begin{equation}
\check{\gammai}(\tau) = \bs{\Phi}(\tau, t_k) \check{\gammai}(t_k),
\end{equation}
where $\check{\gammai}$ is the prior mean, and $\bs{\Phi}(\tau, t_k)$ is the \textit{state transition function} from timestep $t_k$ to timestep $\tau.$

\subsection{GP Prior for SE(3)}
\label{sec:gp_prior}
In $SE(3),$ a physically-motivated GP prior is the following SDE:
\begin{equation}
\label{eq:nonlinear_sde}
\begin{split}
\dot{\mathbf{T}}(t) &= \varpii(t)^\wedge \mathbf{T}(t), \\
\varpidot(t) &= \mathbf{w}^\prime(t),\ \ \ \ \mathbf{w}(t) \sim \mathcal{GP}(\mathbf{0}, \mathbf{Q}_c \delta(t-t')),
\end{split}
\end{equation}
where  $\mathbf{T}(t) = \exp(\xii(t)^\wedge) \in SE(3)$ is the pose with $\xii(t) = \begin{bmatrix}
\bs{\rho}(t)^T & \bs{\phi}(t)^T
\end{bmatrix}^T \in \mathbb{R}^6,$ where $\xii$ is the vector-space representation of the pose. $\bs{\rho}(t) = \begin{bmatrix}
\rho_1(t) & \rho_2(t) & \rho_3(t)
\end{bmatrix}^T,$ $\bs{\phi}(t) = \begin{bmatrix}
\phi_1(t) & \phi_2(t) & \phi_3(t)
\end{bmatrix}^T,$ where $\bs{\rho}$ and $\bs{\phi}$ are the translational and rotational components of $\xii,$ respectively.  $\varpii(t) = \begin{bmatrix}
\bs{\nu}(t)^T &\bs{\omega}(t)^T
\end{bmatrix}^T \in \mathbb{R}^6$ is the body-centric velocity. $(\cdot)^\wedge$ converts $\xii(t) \in \mathbb{R}^6$ into a member of \textit{Lie algebra}, $\mathfrak{se}(3)$ \cite{barfoot2014associating} \cite{barfoot2017state}. The state is
\begin{equation}
\label{eq:wnoa_state}
\mathbf{x}(t) = \{ \mathbf{T}(t), \varpii(t) \}.
\end{equation}
However, it can be seen that the SDE in \eqref{eq:nonlinear_sde} is nonlinear, and therefore cannot be cast into the form of \eqref{eq:lti_sde} and solved efficiently \cite{anderson2015full}. Instead, \cite{anderson2015full} defines a \textit{local} pose variable:
\begin{equation}
\label{eq:local_pose_variable}
\xii_i(t) := \ln(\mathbf{T}(t)\mathbf{T}_i^{-1})^\vee, \ \ t_i \leq t \leq t_{i+1},
\end{equation}
which is a function of the \textit{global} pose variables, $\mathbf{T}(t)$ and $\mathbf{T}_i,$ where $\mathbf{T}_i = \mathbf{T}(t_i).$ For simplicity we use $\mathbf{T}_i$ to denote $\mathbf{T}_{i,0}.$ Here $(\cdot)^\vee$ is the inverse of $(\cdot)^\wedge,$ and converts a member of $\mathfrak{se}(3)$ to $\xii \in \mathbb{R}^6$ \cite{barfoot2014associating} \cite{barfoot2017state}.

Using \textit{local} variables, \cite{anderson2015full} defines a sequence of \textit{local} priors that can be cast into a LTI SDE of the form in \eqref{eq:lti_sde}, with
\begin{equation}
\gammai_i(t) := \begin{bmatrix}
\xii_i(t) \\ \xidot_i(t)
\end{bmatrix}, \ \ \bs{A} = \begin{bmatrix}
\mathbf{0} & \mathbf{1} \\ \mathbf{0} & \mathbf{0}
\end{bmatrix}, \ \ \bs{L} = \begin{bmatrix}
\mathbf{0} \\ \mathbf{1}
\end{bmatrix},
\end{equation}
where $\bs{\gamma}_i(t)$ is defined as the \textit{local} state. Under this formulation, we have white-noise on the second derivative of $\xii_i(t),$ i.e. $\xiddot_i(t) = \mathbf{w}(t).$ Furthermore, we have the following \cite{anderson2015full}:
\begin{equation}
\label{eq:xii_dot}
\xidot_i(t) = \jac(\xii_i(t))^{-1} \varpii(t),
\end{equation}
where $\jac(\xii) \in \mathbb{R}^{6\times6}$ is the \textit{left Jacobian} of $SE(3)$ \cite{barfoot2014associating} \cite{barfoot2017state}.

\subsection{Cost Terms in Optimization}
For our estimator, the negative-log-likelihood objective function consists of prior and measurement cost terms:
\begin{equation}
\label{eq:obj_func}
J = \underbrace{\sum_i J_i}_{\mathrm{prior}} + \underbrace{\sum_j J_j}_{\mathrm{measurement}}\hspace{-3mm}.
\end{equation}
Since our estimator is only for odometry, we do not keep landmarks as part of the state, as in the full STEAM problem. The optimization problem is then
\begin{equation}
\hat{\mathbf{x}} = \argminD_{\mathbf{x}} J(\mathbf{x}),
\end{equation}
where the state $\mathbf{x}$ consists of all trajectory poses and velocities, as defined in \eqref{eq:wnoa_state}. We solve the optimization problem using Gauss-Newton, where each $\mathbf{T}$ and $\varpii$ are updated using an $SE(3)$ perturbation scheme \cite{barfoot2014associating} \cite{anderson2015full}:
\begin{equation}
\label{eq:se3_perturb}
\mathbf{T}_{\mathrm{op},i} \leftarrow \exp(\delta\xii_{i}^{\wedge})\mathbf{T}_{\mathrm{op},i},\ \ \ \ \varpii_{\mathrm{op},i} \leftarrow \varpii_{\mathrm{op},i} + \delta\varpii_i,
\end{equation}
where $(\cdot)_{\mathrm{op}}$ is the operating point. Each prior cost term is
\begin{equation}
\label{eq:prior_cost}
J_i = \frac{1}{2}\mathbf{e}_i^T \mathbf{Q}_i^{-1} \mathbf{e}_i.
\end{equation}
In terms of \textit{local} pose variables, each prior error term is
\begin{equation}
\label{eq:local_error_term}
\mathbf{e}_i = \gammai_i(t_{i+1}) - \hat{\gammai}_i(t_{i+1}) - \bs{\Phi}(t_{i+1}, t_i) (\gammai_i(t_i) - \hat{\gammai}_i(t_i)),
\end{equation}
where the \textit{local} state variables are defined as \cite{anderson2015full}
\begin{equation}
\label{eq:gamma_i}
\gammai_i(t_i) = \begin{bmatrix}
\mathbf{0} \\ \varpii_i
\end{bmatrix}, \ \ \ \ \gammai_i(t_{i+1}) = \begin{bmatrix}
\ln (\mathbf{T}_{i+1,i})^\vee \\ \jac_{i+1,i}^{-1}\varpii_{i+1}
\end{bmatrix},
\end{equation}
with $\mathbf{T}_{i+1,i} := \mathbf{T}_{i+1} \mathbf{T}_{i}^{-1},$ $\jac_{i+1,i} := \jac(\ln(\mathbf{T}_{i+1}\mathbf{T}_i^{-1})^\vee).$
The state transition function can be computed as in \cite{barfoot2017state},
\begin{equation}
\bs{\Phi}(t, t_i) = \exp( \bs{A}\Delta t_i) = \begin{bmatrix}
\mathbf{1} & \Delta t_i\mathbf{1} \\ \mathbf{0} & \mathbf{1}
\end{bmatrix},
\end{equation}
and the inverse covariance matrix is \cite{barfoot2014associating} \cite{anderson2015full}
\begin{equation}
\label{eq:Qi_inv_cv}
\mathbf{Q}_i(t)^{-1} = \begin{bmatrix}
12 \Delta t_i^{-3} \mathbf{Q}_c^{-1} & -6\Delta t_i^{-2} \mathbf{Q}_c^{-1} \\
-6\Delta t_i^{-2} \mathbf{Q}_c^{-1}  & 4\Delta t_i^{-1} \mathbf{Q}_c^{-1} 
\end{bmatrix},
\end{equation}
where $\Delta t_i = t-t_i.$  Using the relationship between \textit{local} and \textit{global} state variables, we can re-write the prior error term in terms of \textit{global} state variables as \cite{anderson2015full}
\begin{equation}
\label{eq:cv_global_error}
\mathbf{e}_i = \begin{bmatrix}
\ln(\mathbf{T}_{i+1}\mathbf{T}_i^{-1})^\vee - (t_{i+1}-t_i)\varpii_i \\
\jac_{i+1,i}^{-1}\varpii_{i+1} - \varpii_i
\end{bmatrix}.
\end{equation}
Each measurement cost term is
\begin{equation}
J_j = \frac{1}{2}\frac{u_j^2}{1+u_j^2},
\end{equation}
where we have chosen the Geman-McClure robust cost \cite{barfoot2017state}. Each $u_j$ is a whitened error norm. Given a point $\mathbf{p}$ measured at time $t = \tau,$ and let $\mathbf{q}$ be its matched point, expressed in the reference frame  $\Vec{\bcf}_0.$ Define a measurement error term:
\vspace{-1mm}
\begin{equation}
\label{eq:meas_error}
\mathbf{g}_j = \mathbf{D} (\mathbf{p} - \mathbf{T}_{\tau} \mathbf{q}),
\vspace{-1mm}
\end{equation}
where $\mathbf{D} \in \mathbb{R}^{3\times4}$ is a projection matrix. If $\mathbf{p}$ lies on a plane, then we formulate a point-to-plane whitened error norm:
\begin{equation}
u_j^{\mathrm{plane}} = \sqrt{\mathbf{g}_j^T(\beta \mathbf{n}\mathbf{n}^T) \mathbf{g}_j},
\end{equation}
where $\mathbf{n}$ is the surface normal of $\mathbf{p},$ and $\beta$ is a scale factor. If $\mathbf{p}$ does not lie on a plane, we formulate a point-to-point whitened error norm:
\begin{equation}
u_j^{\mathrm{point}} = \sqrt{\mathbf{g}_j^T \mathbf{R}_j^{-1} \mathbf{g}_j},
\end{equation}
where $\mathbf{R}_j$ is the associated measurement covariance.

Our lidar odometry algorithm utilizes sliding-window optimization, and runs in an iterative fashion where matched pairs of points are found in each iteration. Please refer to our previous work \cite{tang2018learning} for further details on our odometry pipeline, such as point matching and keypoint selection.

\subsection{Querying the Trajectory}
Our formulation allows us to incorporate measurements at any time along the trajectory, not just at timesteps kept in the state vector as for discrete-time methods. Suppose we have a measurement at $t = \tau$ as in \eqref{eq:meas_error}, and that $t_i < \tau < t_{i+1},$ where $t_i$ and $t_{i+1}$ are knot times in the state. We can interpolate for the state at $\tau$ using results from \cite{anderson2015full}:
\begin{equation}
\label{eq:query_local}
\gammai_i(\tau) = \Lambdai(t)\gammai_i(t_i) + \Omegai(t)\gammai_i(t_{i+1}),
\end{equation}
where $\Lambdai(\tau) \in \mathbb{R}^{12\times12}$ and $\Omegai(\tau) \in \mathbb{R}^{12\times12}$ are \cite{barfoot2014batch}
\begin{equation}
\label{eq:interp_coef}
\begin{split}
\Lambdai(\tau) &= \Phii(\tau, t_i) - \Omegai(\tau) \Phii(t_{i+1}, t_i), \\
\Omegai(\tau) &= \mathbf{Q}_i(\tau) \Phii(t_{i+1}, t)^T \mathbf{Q}_i(\tau)^{-1}.
\end{split}
\end{equation}
Again, using our knowledge on the relationship between \textit{local} and \textit{global} state variables as in \eqref{eq:gamma_i}, we can re-formulate \eqref{eq:query_local} using \textit{global} state variables. While interpolating for the body-centric velocity at an arbitrary time might be of interest to certain applications, for lidar-only odometry we are mainly interested in pose interpolation:
\begin{multline}
\label{eq:query_global}
\mathbf{T}_{\tau} =  \exp \big( \big(\Lambdai_{12}(\tau)\varpii_i + \Omegai_{11}(\tau) \ln(\mathbf{T}_{i+1,i})^\vee \\
+ \Omegai_{12}(\tau)\jac_{i+1,i}^{-1}\varpii_{i+1}\big)^\wedge \big) \mathbf{T}_{i},
\end{multline}
where $\Lambdai_{mn}$ and $\Omegai_{mn}$ are $\mathbb{R}^{6\times6}$ sub-blocks of $\Lambdai$ and $\Omegai.$ This is a principled approach for querying the trajectory that comes directly from standard GP interpolation \cite{rasmussen2004gaussian}. It can be seen that, given a measurement at $t = \tau$ with $t_i < \tau < t_{i+1},$ the result in \eqref{eq:query_global} allows updates to temporally adjacent state variables at $t_i$ and $t_{i+i}$ in the optimization process.

\section{Estimator Bias}
\label{sec:estimator_bias}
\begin{figure}[!htbp]

\centering
\includegraphics[height=1.75in]{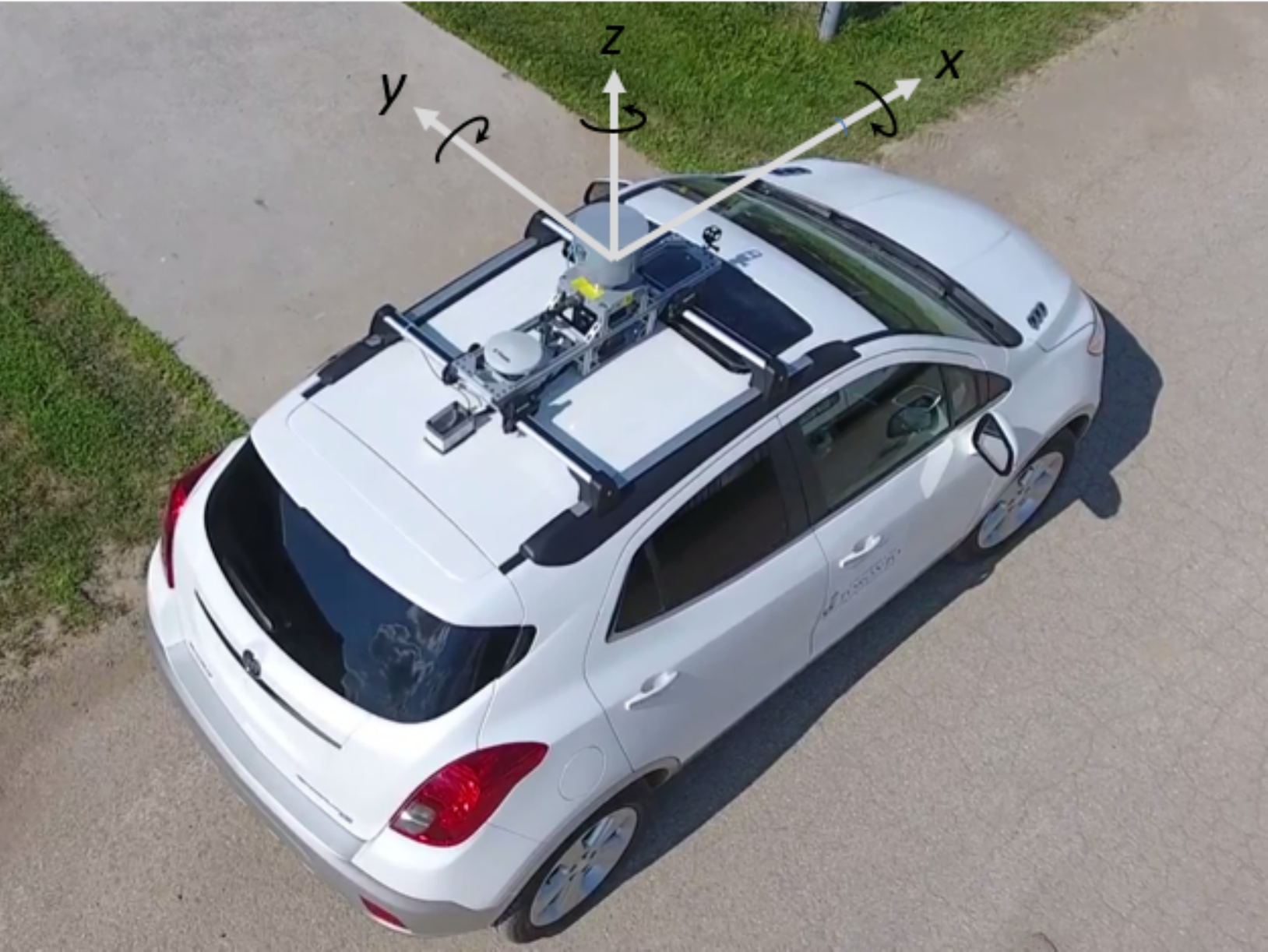}
\captionof{figure}{\footnotesize \label{fig:car_coords} The coordinate system for our estimator. Roll ($\phi_1$), pitch ($\phi_2$), and yaw ($\phi_3$) are rotations about the $x (\rho1),$ $y (\rho_2),$ and $z (\rho_3)$ axes, respectively.}
\vspace{-2mm}
\end{figure}

As shown in Section \ref{sec:experimental_validation} (Figures \ref{fig:kitti_10}, \ref{fig:george_compare}, \ref{fig:leek}), there are noticeable biases in our baseline estimator, particularly in the directions of $z$ ($\rho_3$), roll ($\phi_1$), and pitch ($\phi_2$). See Figure \ref{fig:car_coords} for the coordinate system for our estimator.

There could be many sources that might cause the estimated trajectory to be biased, such as poor sensor calibration, or choosing a measurement covariance that does not reflect the sensor noise characteristics. In this paper, we focus on a very specific source of estimator bias, which results from the motion prior being insufficient to represent the underlying continuous-time trajectory. 

Consider a very simple estimation problem in which a robot, initially stationary at time $t_0$, travels from $\Vec{\bcf}_0$ at $t_0$ to $\Vec{\bcf}_1$ at $t_1$ under constant acceleration. The robot only travels forward, therefore motion only occurs in the $\rho_1$ direction. The robot measures a single point $\mathbf{p}$ at $t_1,$ which is matched against $\mathbf{q}$ measured at $t_0.$ Keeping $\mathbf{T}_0$ and $\varpii_0$ fixed, the state we wish to estimate is
\begin{equation}
\mathbf{x} = \{\mathbf{T}_{1}, \varpii_1\}.
\end{equation}
We can define the following ground truth quantities:
\begin{equation}
\label{eq:update}
\begin{split}
\xii_{\mathrm{gt},1} &=  \begin{bmatrix}
\frac{1}{2}a(t_1-t_0)^2 & 0 & 0 & 0 & 0 & 0
\end{bmatrix}^T, \\
\mathbf{T}_{\mathrm{gt},1} &= \exp(\xii_{\mathrm{gt},1}^{\wedge}),     \\
\varpii_{\mathrm{gt},1} &= \begin{bmatrix}
a(t_1-t_0) & 0 & 0 & 0 & 0 & 0
\end{bmatrix}^T, \\
\end{split}
\end{equation}
where $a$ is the acceleration in $\rho_1.$ We can define the following measurement error equation and the associated measurement Jacobian \cite{barfoot2014associating}:
\begin{equation}
\mathbf{g} = \mathbf{D}( \mathbf{p} - \mathbf{T}_{\mathrm{op},1} \mathbf{q}), \ \ 
\mathbf{G} = -\mathbf{D}(\mathbf{T}_{\mathrm{op},1}\mathbf{q})^\odot,
\end{equation}
where the $(\cdot)^\odot$ operator is defined in \cite{barfoot2014associating}. We also have a WNOA prior error term $\mathbf{e}$ that can be constructed from \eqref{eq:cv_global_error}. For simplicity, we assume $t_1-t_0 = 1,$ $\mathbf{Q} = \mathbf{R} = \mathbf{1},$ and that the measurement is noise-free. Initializing the state variables at ground truth, we have $\mathbf{g} = \mathbf{0}$ as the measurement is noise-free. However, $\mathbf{e}$ is not zero since the motion is not constant-velocity. Performing Gauss-Newton for one iteration using the perturbation scheme in \cite{anderson2015full} (Equation \eqref{eq:se3_perturb}), we have:
\begin{equation}
\label{eq:sym_point_meas}
\begin{split}
\delta\xii_{1} &= \frac{1}{m}\begin{bmatrix}
-\frac{1}{4}a\Big((a^2 - 4x)^2 + 32 (y^2 + z^2 + 1)\Big) \\
ay(a + 4x) \\
az(a + 4x) \\
0 \\
8az \\
-8ay
\end{bmatrix},
\end{split}
\end{equation}

where $m = (a^2 - 4 x)^2 + 16 (y^2 + z^2 + 2),$ and $\mathbf{T}_{\mathrm{op},1}\mathbf{q} = \begin{bmatrix}
x & y & z & 1
\end{bmatrix}^T.$ Equation \eqref{eq:sym_point_meas} shows that our simple problem results in perturbations to degrees of freedom where there is no motion ($\rho_2, \rho_3, \phi_2,$ and $\phi_3$), effectively creating a bias. Moreover, the perturbations to these DOFs depend on $\begin{bmatrix}
x & y & z
\end{bmatrix}^T,$ the Cartesian coordinate of the transformed point, $\mathbf{T}_{\mathrm{op},1}\mathbf{q}.$

We can draw the observation that when the motion prior cannot sufficiently describe the underlying trajectory, such as when the prior mean is constant-velocity but the trajectory is constant-acceleration, then the estimator will be biased in certain degrees of freedom. Particularly, the induced bias is a function of the Cartesian coordinates of points. The bias stems from the optimizer's desire to keep the cost low; the prior cost is made smaller by increasing the measurement cost in the overall objective function \eqref{eq:obj_func}. It can be shown easily that this source of bias will not occur when we use an estimator with a WNOJ prior, as derived in Section \ref{sec:derivation}.

Equation \eqref{eq:sym_point_meas} is computed assuming the robot has forward acceleration. However, a similar case can be made for motion with angular acceleration, such as when initiating a turn. 


\section{White-Noise-On-Jerk Motion Prior}
\label{sec:derivation}
Here we derive a white-noise-on-jerk motion prior. Instead of modelling the acceleration as a zero-mean, white-noise Gaussian process as in the case of a WNOA prior \cite{anderson2015full}, we now explicitly estimate the following state:
\begin{equation}
\mathbf{x}(t) = \{ \mathbf{T}(t), \varpii(t) , \varpidot(t)\}.
\end{equation}
where $\varpidot(t) \in \mathbb{R}^6$ is the body-centric acceleration. 

Extending the idea of \textit{local} pose variables as presented in Section \ref{sec:gp_prior}, we can define a sequence of \textit{local} white-noise-on-jerk priors as a LTI SDE in the form of \eqref{eq:lti_sde}:
\begin{equation}
\gammai_i(t) := \begin{bmatrix}
\xii_i(t) \\ \xidot_i(t) \\ \xiddot_i(t)
\end{bmatrix}, \ \ \bs{A} = \begin{bmatrix}
\mathbf{0} & \mathbf{1} & \mathbf{0} \\
\mathbf{0} & \mathbf{0} & \mathbf{1} \\
\mathbf{0} & \mathbf{0} & \mathbf{0}
\end{bmatrix}, \ \ \bs{L} = \begin{bmatrix}
\mathbf{0} \\ \mathbf{0} \\ \mathbf{1}
\end{bmatrix}.
\end{equation}
We now have white-noise on the third derivative (jerk) of $\xii_i(t),$ $\dddot{\xii}_i(t) = \mathbf{w}(t),$ where $\mathbf{w}(t) \sim \mathcal{GP}(\mathbf{0}, \mathbf{Q}_c \delta(t-t')).$
For the WNOJ prior, the state transition function is now
\begin{equation}
\label{eq:Phi}
\bs{\Phi}(t, t_i) = \exp( \bs{A}\Delta t_i) = \begin{bmatrix}
\mathbf{1} & \Delta t_i\mathbf{1} & \frac{1}{2}\Delta t_i^2\mathbf{1}\\ 
\mathbf{0} & \mathbf{1} & \Delta t_i\mathbf{1}\\
\mathbf{0} & \mathbf{0} & \mathbf{1}
\end{bmatrix},
\end{equation}
and the covariance matrix can be computed as
\begin{equation}
\label{eq:Qi}
\begin{split}
\mathbf{Q}_i(t) = & \int_0^{\Delta t_i} \hspace{-5mm}    \exp(\bs{A}(\Delta t_i-s)) \bs{L}\mathbf{Q}_c\bs{L}^T \exp(\bs{A}(\Delta t_i-s))^T\ ds\\
= & \begin{bmatrix}
\frac{1}{20}\Delta t_i^5\mathbf{Q}_c & \frac{1}{8}\Delta t_i^4\mathbf{Q}_c & \frac{1}{6}\Delta t_i^3\mathbf{Q}_c \\
\frac{1}{8}\Delta t_i^4\mathbf{Q}_c & \frac{1}{3}\Delta t_i^3\mathbf{Q}_c & \frac{1}{2}\Delta t_i^2\mathbf{Q}_c \\
\frac{1}{6}\Delta t_i^3\mathbf{Q}_c & \frac{1}{2}\Delta t_i^2\mathbf{Q}_c & \Delta t_i\mathbf{Q}_c
\end{bmatrix}.
\end{split}
\raisetag{1\baselineskip}
\end{equation}
The inverse covariance matrix is then
\begin{equation}
\label{eq:Qi_inv_ca}
\begin{split}
& \mathbf{Q}_i(t)^{-1}  =\\
& \begin{bmatrix}
720\Delta t_i^{-5}\mathbf{Q}_c^{-1} & -360\Delta t_i^{-4}\mathbf{Q}_c^{-1} & 60\Delta t_i^{-3}\mathbf{Q}_c^{-1} \\
-360\Delta t_i^{-4}\mathbf{Q}_c^{-1} & 192\Delta t_i^{-3}\mathbf{Q}_c^{-1} & -36\Delta t_i^{-2}\mathbf{Q}_c^{-1} \\
60\Delta t_i^{-3}\mathbf{Q}_c^{-1} &- 36\Delta t_i^{-2}\mathbf{Q}_c^{-1} & 9\Delta t_i^{-1}\mathbf{Q}_c^{-1} \\
\end{bmatrix}
\end{split}.
\end{equation}

Figure \ref{fig:traj_samples} shows trajectories sampled from a white-noise-on-jerk prior distribution where the prior mean is constant-acceleration, compared with trajectories sampled from a white-noise-on-acceleration prior distribution where the prior mean is constant-velocity. We argue that the WNOJ prior is more suitable for representing motion with non-zero acceleration trajectory sections, such as in urban driving.

\subsection{Prior Error Term}
In \textit{local} pose variables, the prior error term is the same as in \eqref{eq:local_error_term}. We wish to then express the prior error in terms of $\mathbf{T},$ $\varpii,$ and $\varpidot.$ The relationship between $\xii_i(t)$ and $\xidot_i(t)$ and \textit{global} state variables are shown in Equations \eqref{eq:local_pose_variable} and \eqref{eq:xii_dot}. To express $\xiddot_i(t)$ in terms of \textit{global} state variables, we have
\begin{equation}
\begin{split}
\ddot{\bs{\xi}}_i(t) &= \frac{d}{dt}(\dot{\bs{\xi}}_i(t)) = \frac{d}{dt}(\bs{\mathcal{J}}(\bs{\xi}_i(t))^{-1}\bs{\varpi}(t)) \\
&= \frac{d}{dt}(\bs{\mathcal{J}}(\bs{\xi}_i(t))^{-1})\bs{\varpi}(t) + \bs{\mathcal{J}}(\bs{\xi}_i(t))^{-1} \varpidot(t).
\end{split}
\end{equation}
We can write the inverse left Jacobian of $SE(3)$ as a power-series expansion \cite{barfoot2017state}:
\begin{equation}
\jac(\xii)^{-1} = \sum_{n=0}^\infty \frac{B_n}{n!}(\xii^\curlywedge)^n = B_0 + \frac{B_1}{1!}\xii^\curlywedge + \frac{B_2}{2!}(\xii^\curlywedge)^2 + \cdots,
\end{equation}
where the coefficients, $B_n,$ are the \textit{Bernoulli numbers}. The operator $(\cdot)^\curlywedge$ is defined as \cite{barfoot2014associating} \cite{barfoot2017state}
\begin{equation}
\xii^\curlywedge = \begin{bmatrix}
\bs{\rho} \\ \bs{\phi}
\end{bmatrix}^\curlywedge = \begin{bmatrix}
\bs{\phi}^\wedge & \bs{\rho}^\wedge \\ \mathbf{0} & \bs{\phi}^\wedge
\end{bmatrix}.
\end{equation}
It can be shown easily that $\frac{d}{dt}(\xii^\curlywedge) = \xidot^\curlywedge,$ therefore
\begin{multline}
\label{eq:correct}
\begin{split}
\frac{d}{dt}(\bs{\mathcal{J}}^{-1}) = \frac{B_1}{1!}\xidot^\curlywedge + \frac{B_2}{2!}(\xidot^\curlywedge \xii^\curlywedge + \xii^\curlywedge \xidot^\curlywedge) + \\
\frac{B_3}{3!}\Big( \xidot^\curlywedge (\xii^\curlywedge)^2  + \xii^\curlywedge \xidot^\curlywedge \xii^\curlywedge + (\xii^\curlywedge)^2\xidot^\curlywedge \Big) + \cdots.
\end{split}
\end{multline}
As it turns out, we cannot express $\frac{d}{dt}(\bs{\mathcal{J}}^{-1})$ analytically in terms of $\bs{\mathcal{J}}^{-1}$ or $\bs{\mathcal{J}},$ which are familiar terms with which to work. We instead resort to the first-order approximation \cite{barfoot2017state} that
\begin{equation}
\label{eq:jac_approx}
\jac(\xii)^{-1} \approx \mathbf{1} - \frac{1}{2}\xii^\curlywedge.
\end{equation}
Our approximation is reasonable as long as $\xii$ is small, which it will be in our case. Under this approximation, we have
\begin{equation}
\frac{d}{dt}(\jac(\xii)^{-1}) \approx -\frac{1}{2}\xidot^\curlywedge,
\end{equation}
and finally
\begin{equation}
\begin{split}
\ddot{\bs{\xi}}_i(t) &= \frac{d}{dt}(\bs{\mathcal{J}}(\bs{\xi}_i(t))^{-1})\bs{\varpi}(t) + \bs{\mathcal{J}}(\bs{\xi}_i(t))^{-1} \varpidot(t) \\
&\approx -\frac{1}{2} \xidot_i(t)^\curlywedge \varpii(t) + \jac(\xii_i(t))^{-1}\varpidot(t) \\
&= -\frac{1}{2}(\jac(\xii_i(t))^{-1}\varpii(t))^\curlywedge\varpii(t) + \jac(\xii_i(t))^{-1}\varpidot(t).
\end{split}
\raisetag{2\baselineskip}
\end{equation}
The \textit{local} state variables can then be written as
\begin{equation}
\label{eq:wnoj_local_global}
\begin{split}
\gammai_i(t_i) &= \begin{bmatrix}
\mathbf{0} \\ \varpii_i \\ \varpidot_i
\end{bmatrix}, \\
\gammai_i(t_{i+1}) &= \begin{bmatrix}
\ln(\mathbf{T}_{i+1,i})^\vee \\ \jac_{i+1,i}^{-1}\varpii_{i+1} \\ -\frac{1}{2}(\jac_{i+1,i}^{-1}\varpii_{i+1})^\curlywedge\varpii_{i+1} + \jac_{i+1,i}^{-1}\varpidot_{i+1}
\end{bmatrix}, \\
\end{split}
\raisetag{3.5\baselineskip}
\end{equation}
where we have made use of the identity $\mathbf{x}^\curlywedge \mathbf{x} = \mathbf{0}$ \cite{barfoot2017state}.

In terms of \textit{global} state variables, the prior error term is
\begin{equation}
\label{eq:ca_global_error}
\mathbf{e}_i = \begin{bmatrix}
\ln(\mathbf{T}_{i+1,i})^\vee - (t_{i+1}-t_i)\varpii_i - \frac{1}{2}(t_{i+1}-t_i)^2\varpidot_i \\
\jac_{i+1,i}^{-1}\varpii_{i+1} - \varpii_i - (t_{i+1}-t_i)\varpidot_i \\
-\frac{1}{2}(\jac_{i+1,i}^{-1}\varpii_{i+1})^\curlywedge\varpii_{i+1} + \jac_{i+1,i}^{-1}\varpidot_{i+1}-\varpidot_i
\end{bmatrix}.
\end{equation}

Suppose we assume the trajectory has zero acceleration (which is assumed by a prior mean that is constant-velocity), $\varpidot_i = \varpidot_{i+1} = \mathbf{0},$ and also make the assumption that $\jac_{i+1,i}^{-1}\varpii_{i+1} \approx \varpii_{i+1}.$ In this case, the last component in \eqref{eq:ca_global_error} becomes zero, and the first two components become identical to the WNOA prior as in \eqref{eq:cv_global_error}; we have essentially recovered the prior error equation for the WNOA prior. 

\subsection{Querying the Trajectory}
We start from the same interpolation equation using \textit{local} state variables \eqref{eq:query_local}. For the WNOJ prior, the interpolation coefficients $\Lambdai(\tau)$ and $\Omegai(\tau)$ can be computed from \eqref{eq:interp_coef}, using $\Phii \in \mathbb{R}^{18\times18}$ and 
$\mathbf{Q}_i \in \mathbb{R}^{18\times18}$ from \eqref{eq:Phi} and \eqref{eq:Qi}.

 Substituting with \textit{global} state variables for the WNOJ prior using \eqref{eq:wnoj_local_global}, the pose interpolation equation is
\begin{multline}
\label{eq:ca_pose_interp_global}
\mathbf{T}_{\tau} = \exp\big( \big( \Lambdai_{12}(\tau)\varpii_i + \Lambdai_{13}(\tau)\varpidot_i + \Omegai_{11}(\tau) \ln(\mathbf{T}_{i+1,i})^\vee \\
 + \Omegai_{12}(\tau)\jac_{i+1,i}^{-1}\varpii_{i+1} + \Omegai_{13}(\tau)(-\frac{1}{2}(\jac_{i+1,i}^{-1}\varpii_{i+1})^\curlywedge\varpii_{i+1} \\
 + \jac_{i+1,i}^{-1}\varpidot_{i+1})\big)^\wedge\big) \mathbf{T}_{i},
\end{multline}
\hspace{-3mm}
where $t_i < \tau < t_{i+1}.$ $\Lambdai_{mn}$ and $\Omegai_{mn}$ are $\mathbb{R}^{6\times6}$ sub-blocks of $\Lambdai(\tau) \in \mathbb{R}^{18\times18}$ and $\Omegai(\tau) \in \mathbb{R}^{18\times18}.$ 

Again, if we assume that $\varpidot_i = \varpidot_{i+1} = \mathbf{0},$ and $\jac_{i+1,i}^{-1}\varpii_{i+1} \approx \varpii_{i+1},$ the terms with coefficients $\Lambdai_{13}$ and $\Omegai_{13}$ become zeros. Similar to the case with the prior error term, we can essentially recover the pose interpolation equation for the WNOA prior as in \eqref{eq:query_global}.

\section{Experimental Validation}
\label{sec:experimental_validation}
To evaluate the white-noise-on-jerk prior we derived, we formulated a variation of our continuous-time lidar odometry estimator that employs the WNOJ prior. The new estimator is evaluated on various Velodyne lidar datasets, and the odometry errors are compared against the baseline estimator presented in Section \ref{sec:ct_estimator}, which employs a WNOA prior. To ensure a fair comparison, all other aspects of STEAM such as constructing measurement terms, and the other components of lidar odometry such as the point matching method, are kept the same. Any differences between the two estimators arise solely from their different motion priors. Our evaluations are odometric, and we make no attempt to use mapping or loop-closure to reduce the estimation error.

The power spectral density matrix, $\mathbf{Q}_c,$ is the only hyperparameter for our lidar odometry algorithm. In practice, $\mathbf{Q}_c$ is designed to be a diagonal matrix, where tuning is done by choosing the diagonal elements from a list of candidates  based on our prior knowledge about the trajectory. While we do not make use of the nonholomic constraint in our estimator, we penalize acceleration or jerk in each DOF differently, by scaling each diagonal element of $\mathbf{Q}_c$ relative to the others. For both types of motion priors, we tuned $\mathbf{Q}_c$ to achieve the best performance on the training set (sequences 0 to 10) of the KITTI odometry benchmark \cite{Geiger2012CVPR}. $\mathbf{Q}_c$ was then kept the same when evaluating on all other datasets.

\subsection{KITTI Odometry Benchmark}
Sequences 0 to 10 are the training sequences of KITTI. Sequences 11 to 21 are the test sequences, where the ground truths are not publicly available. The KITTI benchmark evaluates percentage translation errors across path segments of lengths $100, 200, \dots, 800$ meters, and an average over all path segments is computed \footnote{ \vspace{0mm} \footnotesize The evaluation metric is provided in KITTI's \href{https://s3.eu-central-1.amazonaws.com/avg-kitti/devkit_odometry.zip}{development kit.}}. A total error averaged over path segments evaluated for all sequences is also reported.

\begin{figure}[!htbp]
  \centering
  \includegraphics[height=2.25in]{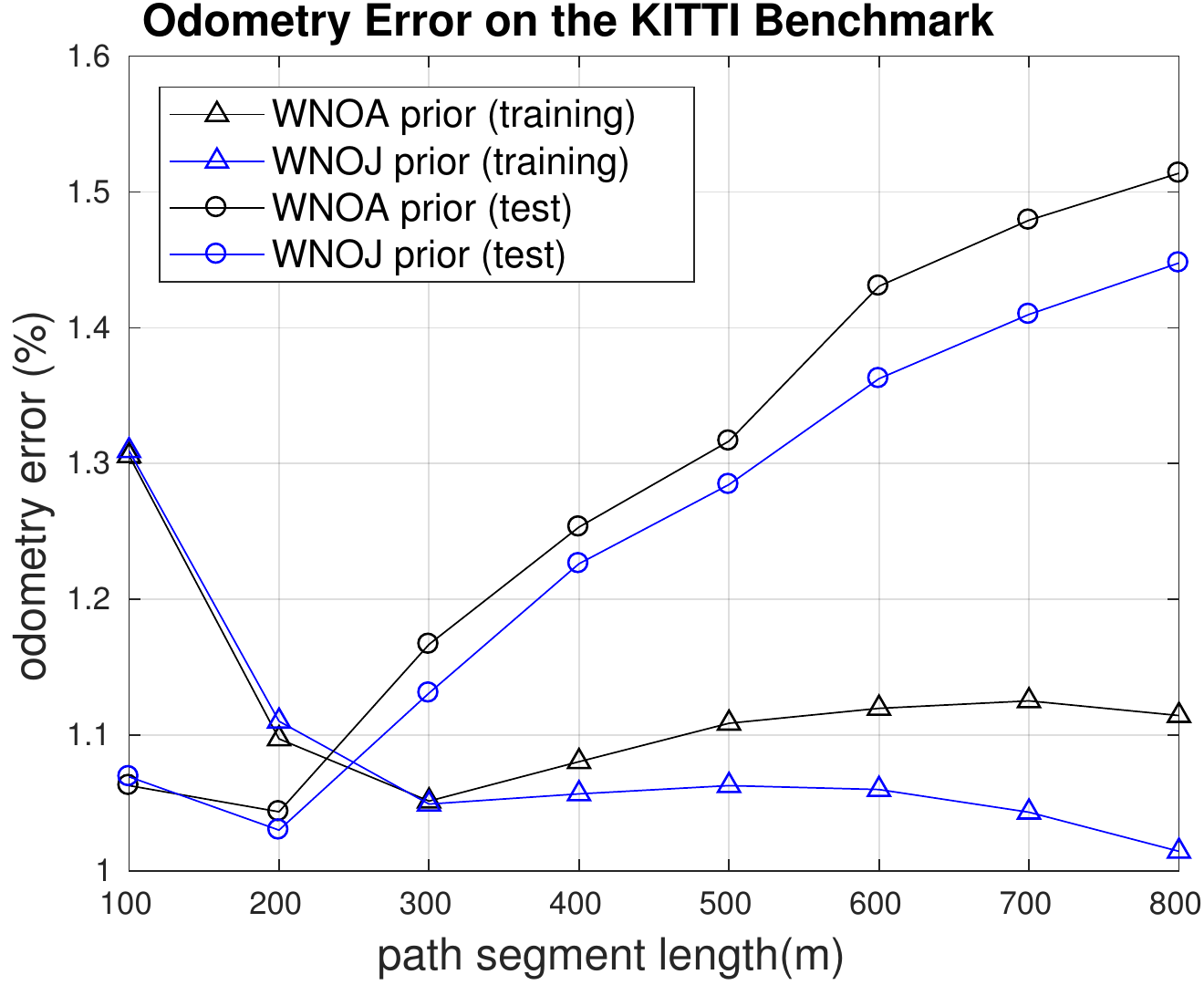}
  \caption{\footnotesize Odometry error for the baseline estimator which employs the WNOA prior, and for the new estimator which employs the WNOJ prior.}
  \label{fig:kitti_error}
\end{figure}

\begin{figure}[!htbp]
  \vspace{-2mm}
  \centering
  \includegraphics[height=2in,clip]{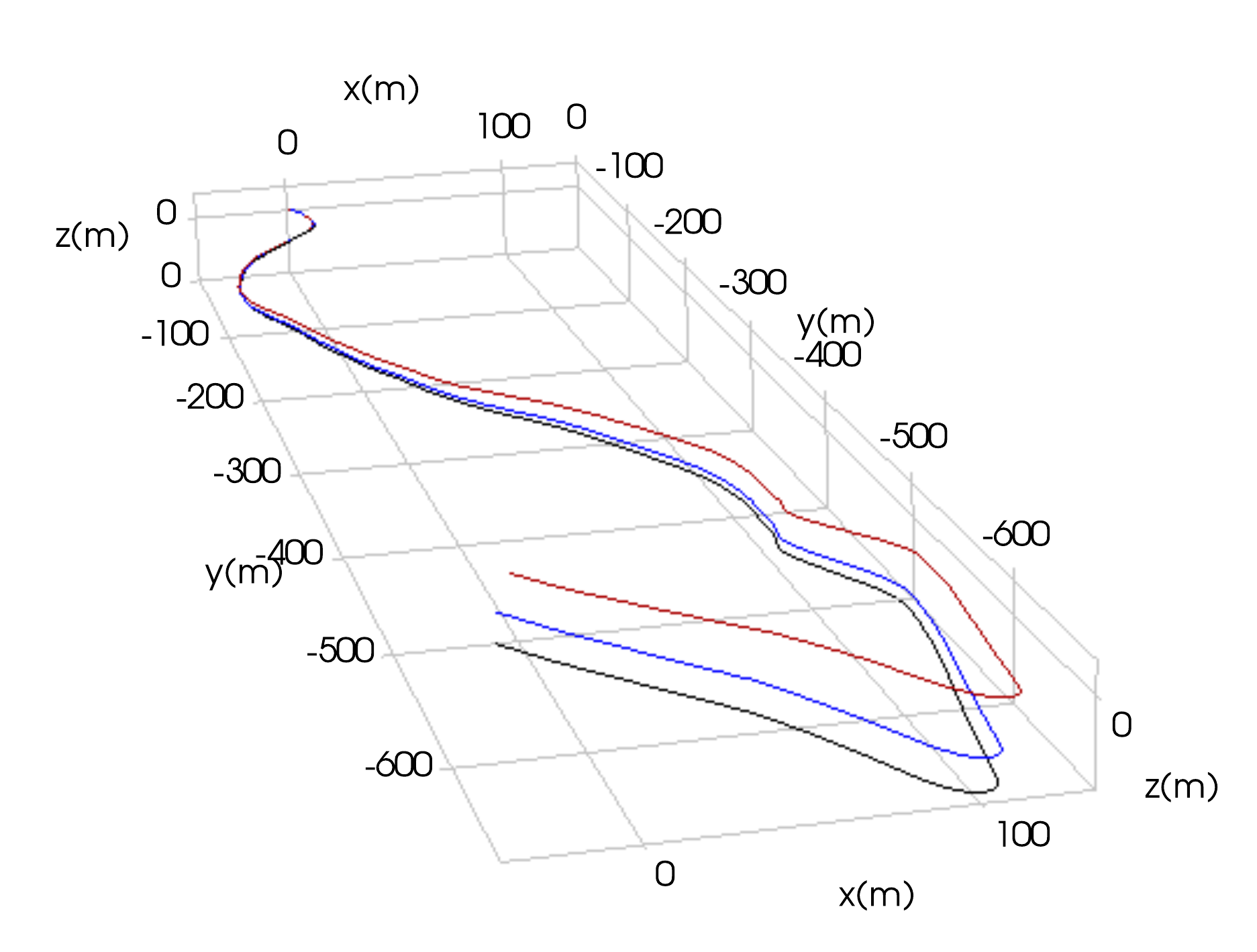}
  \caption{\footnotesize 3D plots of odometry estimates for sequence 10: baseline estimator using WNOA prior (black) vs. new estimator using WNOJ prior (blue) when compared against ground truth (red).}
  \label{fig:kitti_10}
  \vspace{-2mm}
\end{figure}

The baseline estimator that employs a WNOA prior achieved an overall error of $1.13\%$ on the training set, and $1.26\%$ on the test set. Our new estimator that employs a WNOJ prior achieved an overall error of $1.10\%$ on the training set, and $1.22\%$ on the test set. A detailed break-down of the error for various path segment lengths is presented in Figure \ref{fig:kitti_error}. The new estimator using WNOJ prior outperforms the baseline estimator for almost all path segment lengths. Figure \ref{fig:kitti_10} shows a sequence where the odometry biases are noticeably reduced when we use the WNOJ prior.

Our baseline estimator, submitted as \textit{STEAM-L}, currently ranks at $\#11$ on the KITTI leader board among lidar-only methods, while our new estimator with the WNOJ prior, \textit{STEAM-L WNOJ}, ranks at $\#10.$ Both estimators are odometric and make no use of loop-closure. By choosing a motion prior that we believe is more representative of real-world vehicle trajectories, we achieved consistent improvements to our baseline method, which is already fairly accurate.

The lidar point-clouds from the KITTI odometry benchmark were post-processed by the dataset authors to compensate motion distortion. As a result, all points in a sensor revolution can be treated as being measured at exactly the same time, and we do not need to rely on the motion prior for interpolating the pose as in Equations \eqref{eq:query_global} and \eqref{eq:ca_pose_interp_global}. The prior cost terms \eqref{eq:prior_cost}, however, are still used to smooth the trajectory. Nevertheless, for undistorted data such as in the KITTI dataset, it is not necessary to use a continuous-time estimation framework. Even though the new estimator with the WNOJ prior outperformed our baseline estimator on the KITTI benchmark, we argue that datasets with motion-distorted point-clouds are more suitable for comparing continuous-time methods.

\begin{figure*}[!htbp]
\centering
\begin{minipage}{.5\textwidth}
  \centering
  \adjincludegraphics[height=1.6in,clip]{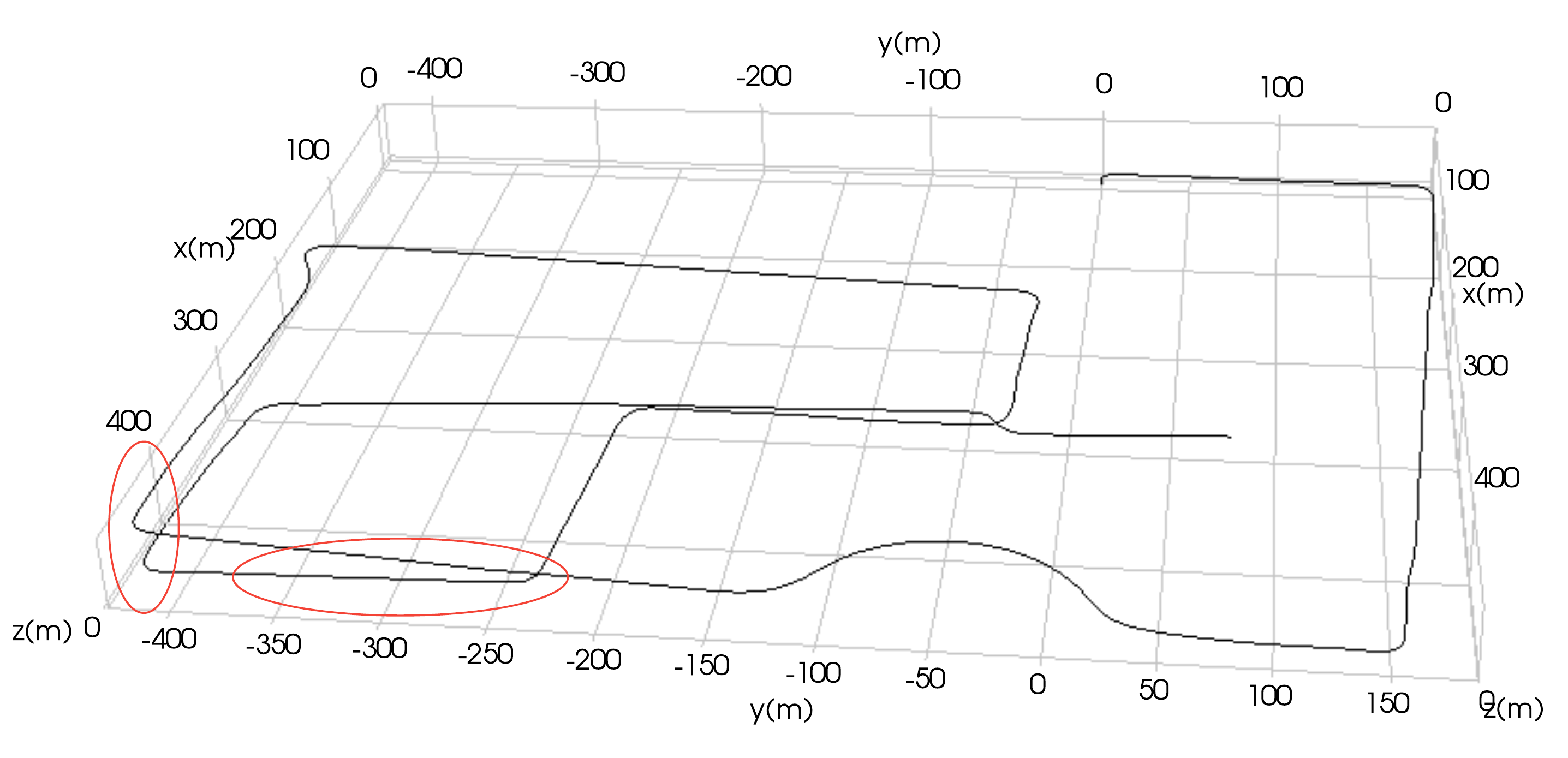}
  \label{fig:0_before}
\end{minipage}%
\begin{minipage}{.5\textwidth}
  \centering
    \adjincludegraphics[height=1.6in,,clip]{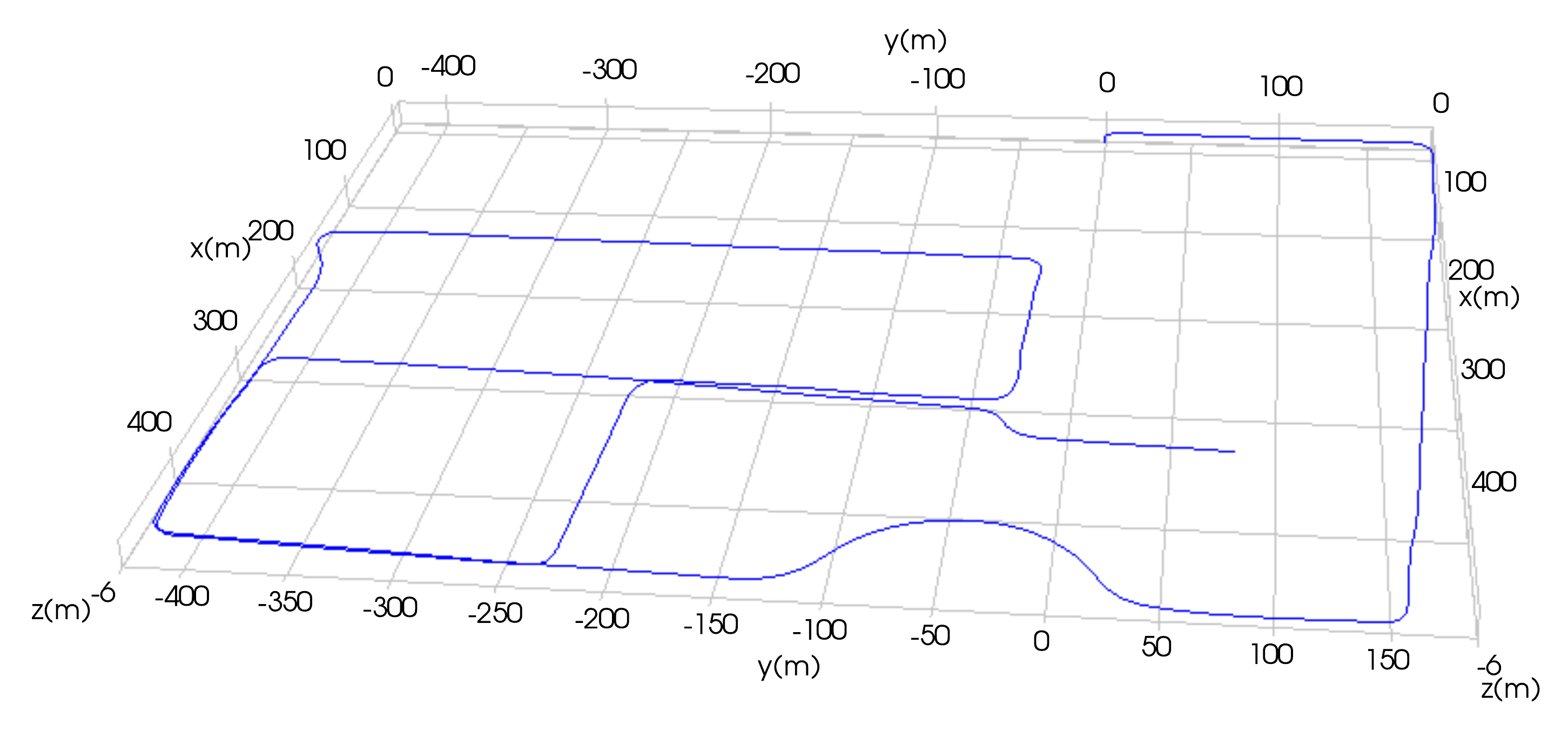}
  \label{fig:0_after}
\end{minipage}
\begin{minipage}{.5\textwidth}
  \centering
  \adjincludegraphics[height=1.34in,clip]{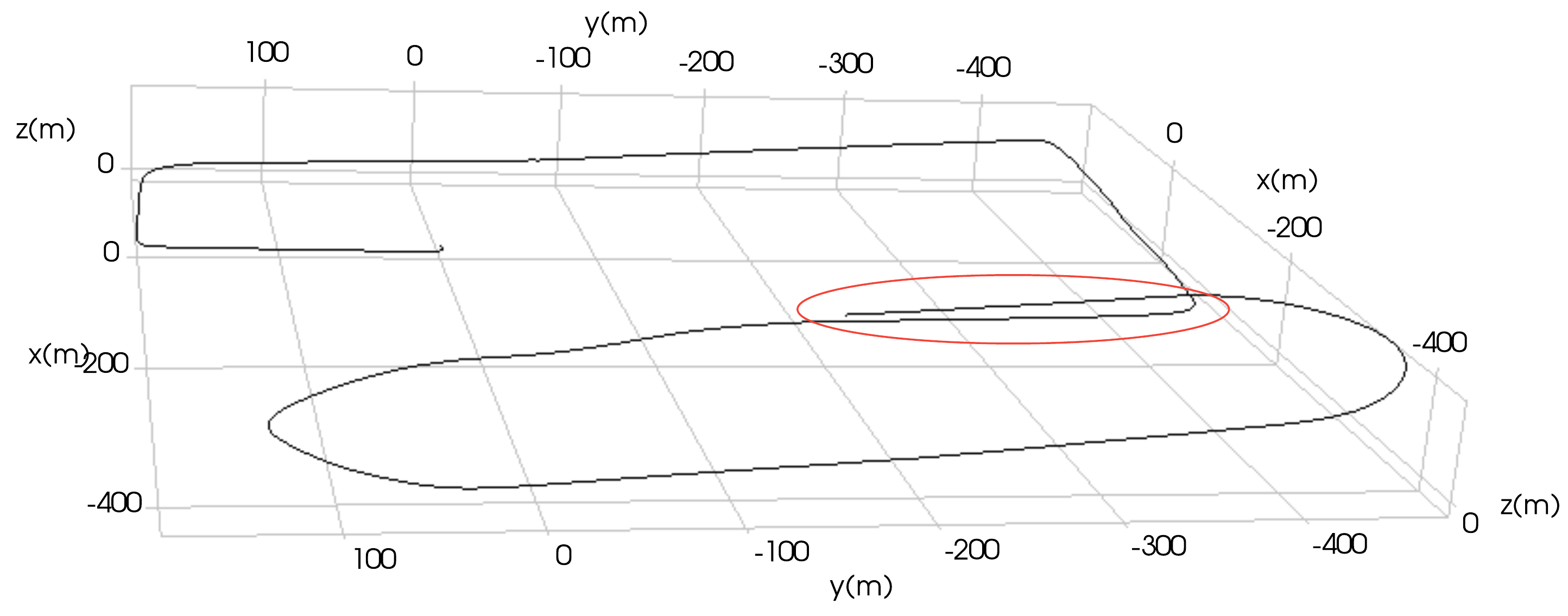}
  \label{fig:0_before}
\end{minipage}%
\begin{minipage}{.5\textwidth}
  \centering
    \adjincludegraphics[height=1.34in,,clip]{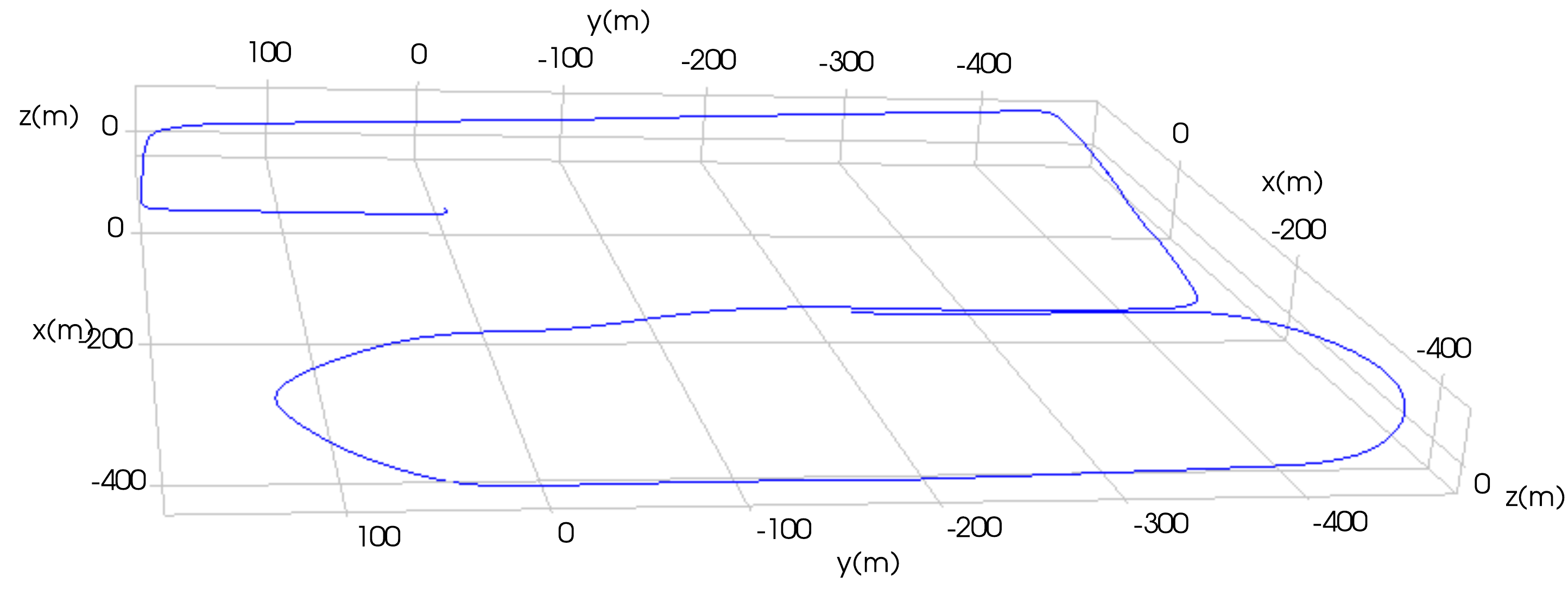}
  \label{fig:0_after}
\end{minipage}
\vspace{-5mm}
\captionof{figure}{\footnotesize \label{fig:george_compare} 3D plots of odometry estimates for sequence 0 (top) and sequence 4 (bottom) of the University of Toronto dataset from the same perspective. Black is odometry using the baseline estimator with WNOA prior, and blue is the new estimator using WNOJ prior. Left: due to biases, odometry does not overlap when the vehicle travels back to a path it has been before (circled). Right: the biases are significantly reduced when using WNOJ prior.}
\end{figure*}

\subsection{University of Toronto Dataset}
\begin{figure}[!htbp]
  \centering
  \includegraphics[height=1.8in]{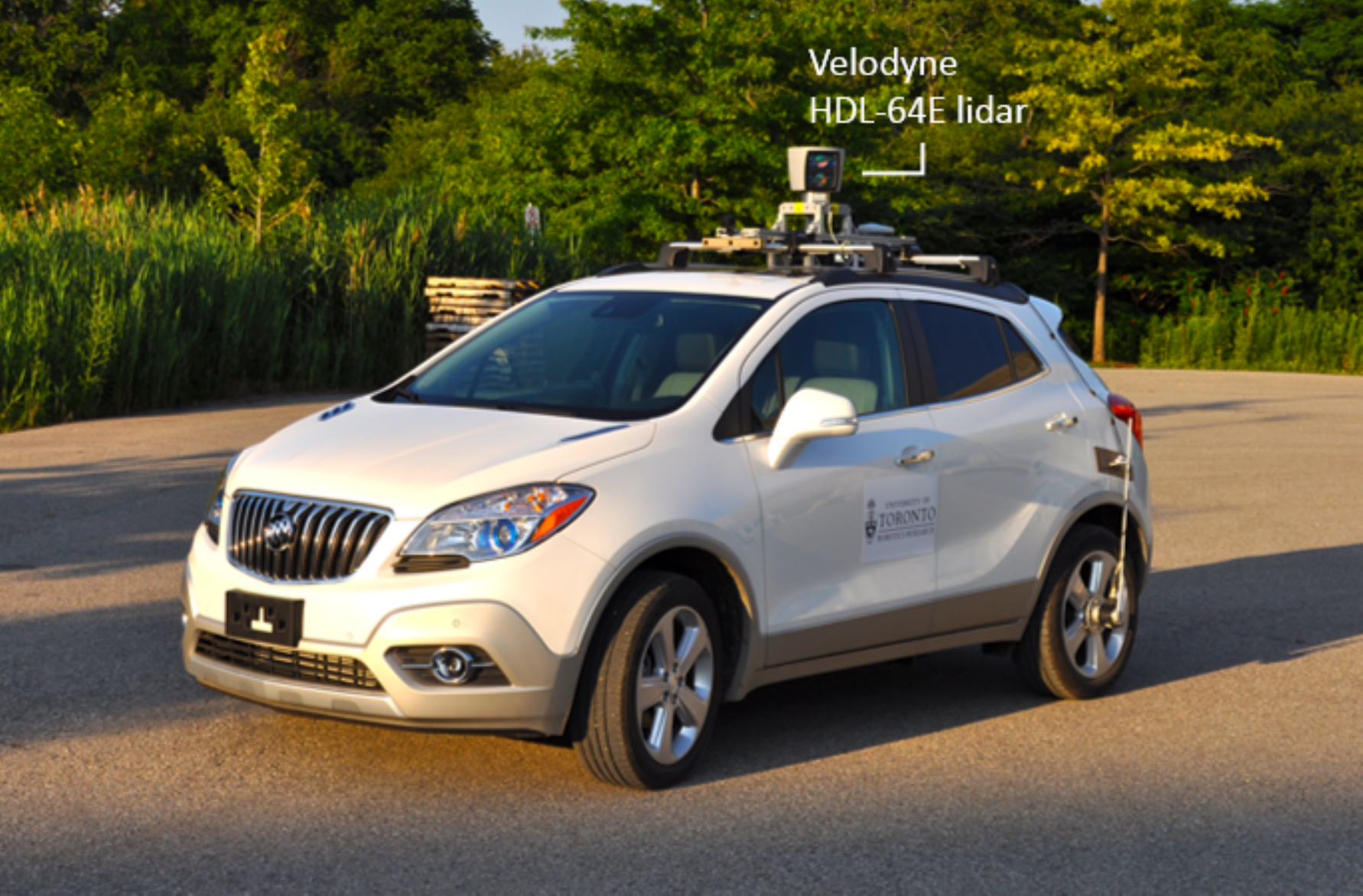}
  \caption{\footnotesize The Buick test vehicle used, equipped with a Velodyne HDL-64E lidar, and an Applanix POS-LV 210 system for ground truth.}
  \label{fig:car}
  \vspace{-3mm}
\end{figure}

A dataset was collected by our test vehicle (Figure \ref{fig:car}) along different routes around University of Toronto (U of T). This resulted in 9 sequences of Velodyne data where each is at least 1.7\,km in distance. 6-DOF ground truth is available via an on-board Applanix positioning and orientation system (POS). For consistency, to evaluate for odometry errors we use the same method as the KITTI benchmark, where translational errors are evaluated across path segments of lengths $100, 200, \dots, 800$ meters. This is a motion-distorted lidar dataset, as we do not employ external sensors or ground truth to compensate the point-clouds. We rely solely on the continuous-time estimator for handling motion distortion.

\begin{table}[]
\caption{\footnotesize \label{tab:uoft_results} Odometry errors for the baseline estimator using WNOA prior and new estimator using WNOJ prior, evaluated on the U of T dataset.}

\begin{tabular}{cccc}
\begin{tabular}[c]{@{}c@{}}Sequence\\ no.\end{tabular} & \begin{tabular}[c]{@{}c@{}}Distance\\ (km)\end{tabular} & \begin{tabular}[c]{@{}c@{}}Baseline estimator\\ with WNOA prior (\%)\end{tabular} & \begin{tabular}[c]{@{}c@{}}New estimator\\ with WNOJ prior (\%)\end{tabular} \\ \hline
0                                                      & 3.34                                                    & 1.5326                                                                            & \textbf{1.2663}                                                              \\
1                                                      & 2.21                                                    & 1.3706                                                                            & \textbf{1.2797}                                                              \\
2                                                      & 3.04                                                    & 1.3967                                                                            & \textbf{1.3485}                                                              \\
3                                                      & 2.91                                                    & 1.7980                                                                             & \textbf{1.5844}                                                              \\
4                                                      & 2.99                                                    & 1.6100                                                                              & \textbf{1.4307}                                                              \\
5                                                      & 1.71                                                    & 2.4319                                                                            & \textbf{2.1696}                                                              \\
6                                                      & 3.48                                                    & 2.1322                                                                            & \textbf{2.054}                                                               \\
7                                                      & 3.04                                                    & 1.2932                                                                            & \textbf{1.2122}                                                              \\
8                                                      & 2.92                                                    & 1.6988                                                                            & \textbf{1.5327}                                                              \\
overall                                                & 25.63                                                   & 1.6736                                                                            & \textbf{1.5235}                                                             
\end{tabular}
\vspace{-5mm}
\end{table}

The U of T dataset features driving in urban scenes where the vehicle's speed is generally under $50\mathrm{km/h}$. However, the vehicle needs to constantly slow down for traffic, or take a turn at an intersection. Since the vehicle's trajectory contains many sections where the velocity changes consistently, this dataset is much more suitable for motion estimation using a WNOJ prior, than a WNOA prior.

The results for the baseline estimator using WNOA prior and the new estimator using WNOJ are compared in Table \ref{tab:uoft_results}. The errors are higher than in the KITTI dataset, mostly because the point-clouds are distorted. The WNOJ prior resulted in smaller error for all sequences, and an overall of $9\%$ error reduction. Figure \ref{fig:george_compare} shows comparison plots of the estimated trajectory using the WNOA prior and WNOJ prior for two sequences from the U of T dataset. The estimated trajectory using a WNOJ prior is significantly more accurate.

Again, since the lidar data are motion-distorted, we need to interpolate the pose for each point measurement. We argue that the WNOJ prior \eqref{eq:ca_pose_interp_global} offers a more suitable interpolation scheme than the WNOA prior \eqref{eq:query_global}. Results for the U of T dataset achieved a greater reduction in error from using the WNOJ prior than the KITTI dataset, which makes no use of the interpolation scheme.

\subsection{Richmond Hill Dataset}
\begin{figure}[!htbp]
\vspace{-2mm}
  \centering
  \includegraphics[height=1.5in,clip]{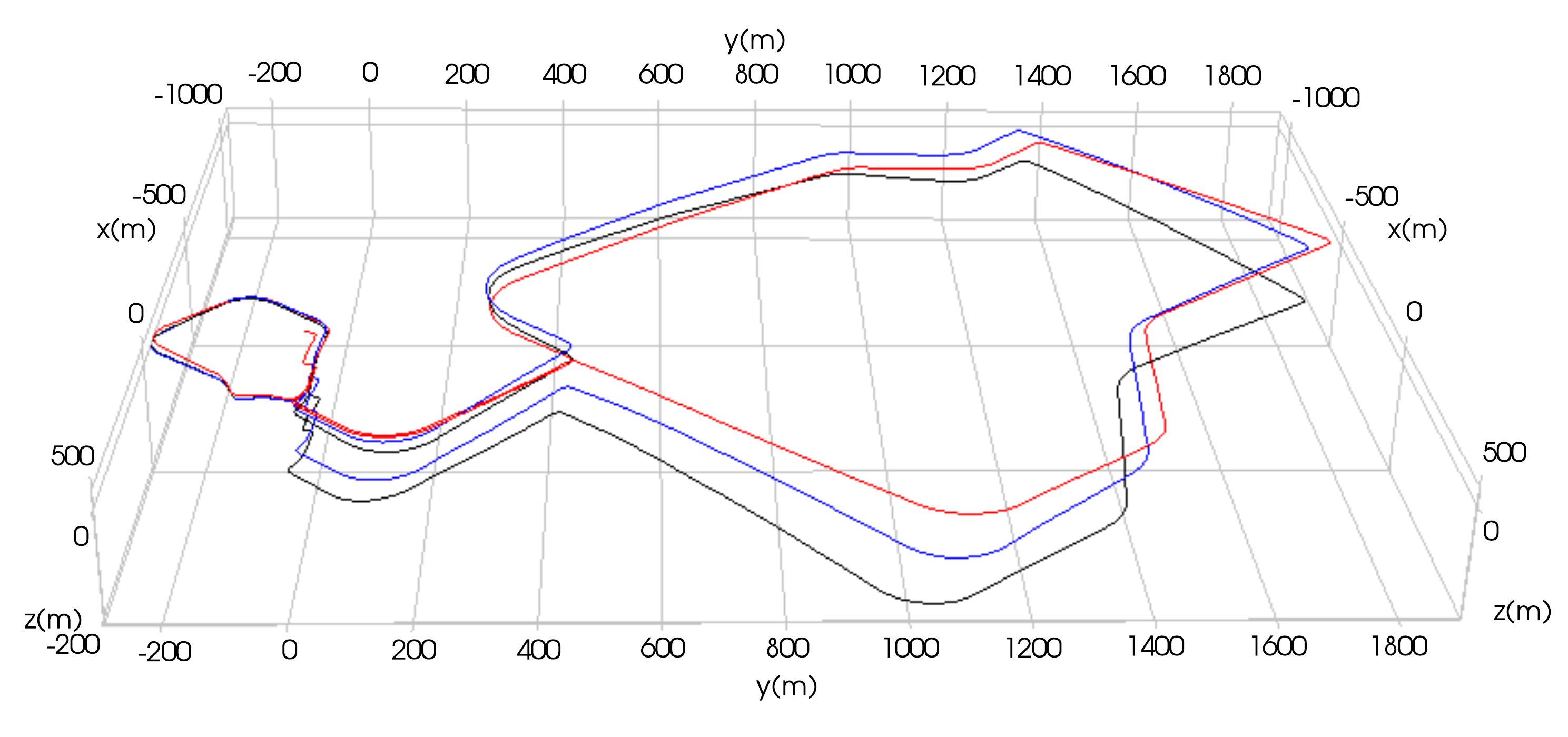}
  \caption{\footnotesize 3D plots of odometry estimates for sequence 1 of the Richmond Hill dataset: baseline estimator using WNOA prior (black) vs. new estimator using WNOJ prior (blue) when compared against ground truth (red).}
  \label{fig:leek}
\vspace{-2mm}
\end{figure}

A dataset was collected in the city of Richmond Hill, North of Toronto, using our test vehicle. This resulted in three long sequences more than $60\mathrm{km}$ in total. The Richmond Hill dataset features driving in suburban areas, and on highways, which contain less useful geometry and structure. Moreover, the test vehicle was driving more than $90\mathrm{km/h}$ on highways, making this a highly challenging dataset. Similar to the U of T dataset, the point-clouds are motion-distorted.

\begin{table}[]
\vspace{3mm}
\caption{\footnotesize \label{tab:rh_results} Errors for the baseline estimator using WNOA prior and new estimator using WNOJ prior, evaluated on the Richmond Hill dataset.}
\begin{tabular}{lccc}
\begin{tabular}[c]{@{}c@{}}Sequence\\ no.\end{tabular} & \begin{tabular}[c]{@{}c@{}}Distance\\ (km)\end{tabular} & \begin{tabular}[c]{@{}c@{}}Baseline estimator\\ with WNOA \\ prior (\%)\end{tabular} & \begin{tabular}[c]{@{}c@{}}New estimator\\ with WNOJ \\ prior (\%)\end{tabular} \\ \hline
0 (suburban)                                            & 17.91                                                   & 1.5887                                                                            & \textbf{1.5094}                                                              \\
1 (urban)                                               & 7.49                                                    & 2.3229                                                                            & \textbf{1.9363}                                                              \\
2 (highway)                                             & 35.01                                                   & 2.3449                                                                            & \textbf{2.1627}                                                              \\
overall                                                & 60.41                                                   & 2.1180                                                                            & \textbf{1.9409}                                                             
\end{tabular}
\vspace{-5mm}
\end{table}
The results are summarized in Table \ref{tab:rh_results}. The new estimator using the WNOJ prior outperformed the baseline estimator for all sequences, resulting in a reduction of the overall error by $8.4\%$. Figure \ref{fig:leek} shows a comparison plot of odometry estimates using the WNOJ prior and the WNOA prior, where the odometry from the new estimator is noticeably less biased than the odometry from the baseline estimator, when compared against ground truth.

\subsection{Remarks}
We found that the new estimator using the WNOJ prior is more sensitive to $\mathbf{Q}_c$ than the baseline estimator using the WNOA prior. This is noticeable because of the difference on the coefficient of each term in the inverse covariance matrix, $\mathbf{Q}_i(t)^{-1},$ between the WNOJ prior \eqref{eq:Qi_inv_ca} and the WNOA prior \eqref{eq:Qi_inv_cv}. Despite this, we achieved an improvement on all datasets using $\mathbf{Q}_c$ tuned on the KITTI training set alone. We plan on releasing our datasets for public use in the future.

The cost of STEAM increases linearly with the number of state variables,  therefore optimizing for \eqref{eq:obj_func} is now $50\%$ more expensive when using the WNOJ prior. The overall increase in runtime is smaller, since the cost for the other components of the estimator, such as point matching, stays invariant.

\section{Conclusion and Future Work}
\label{sec:conclusion}

In this paper, we showed that in continuous-time trajectory estimation, a source of estimator bias can arise when the motion prior cannot sufficiently represent the underlying trajectory. The main contribution of this paper is the derivation of a white-noise-on-jerk motion prior for continuous-time trajectory estimation on $SE(3).$ We showed that the new prior outperforms the existing white-noise-on-acceleration prior employed by STEAM on various lidar datasets, both with and without motion distortion.

Our new formulation of STEAM using the WNOJ prior now has accelerations in the state. Therefore, an extension would be to formulate an estimator that incorporates acceleration measurements from an inertial measurement unit (IMU) directly, rather than pre-integrating to a fixed timestep as is done in many existing inertial estimators.


\bibliographystyle{IEEEtran}
\bibliography{bibi}

\end{document}